%% file: my_paper.tex
\documentclass{article}
\usepackage{iclr2025_conference,times}
\pdfoutput=1
\input{math_commands}

\input{my_packages}

\title{Investigating the Impact of Model Complexity\\ in Large Language Models}

\author{Jing Luo\textsuperscript{1}
\authorskip Huiyuan Wang\textsuperscript{2}
\authorskip Weiran Huang\textsuperscript{3}
\\
\textsuperscript{1}School of Mathematics and Statistics, Shandong University \\
\textsuperscript{2}Perelman School of Medicine, University of Pennsylvania \\
\textsuperscript{3}MIFA Lab, Qing Yuan Research Institute, SEIEE, Shanghai Jiao Tong University 
}

\iclrfinalcopy
\begin{document}

\maketitle

\input{sections/abstract}
\input{sections/intro}
\input{sections/formulation}

\input{sections/analysis}

\input{sections/simulation}
\input{sections/conclusion}
\clearpage
\bibliography{ref}

\clearpage
\input{sections/appendix}

\end{document}

%% file: math_commands.tex
\usepackage{amsmath,amsfonts,bm}

\def\eqref#1{equation~\ref{#1}}

\def\1{\bm{1}}

\DeclareMathAlphabet{\mathsfit}{\encodingdefault}{\sfdefault}{m}{sl}
\SetMathAlphabet{\mathsfit}{bold}{\encodingdefault}{\sfdefault}{bx}{n}



%% file: sections/abstract.tex
\begin{abstract}
    Large Language Models (LLMs) based on the pre-trained fine-tuning paradigm have become pivotal in solving natural language processing tasks, consistently achieving state-of-the-art performance. Nevertheless, the theoretical understanding of how model complexity influences fine-tuning performance remains challenging and has not been well explored yet. In this paper, we focus on autoregressive LLMs and propose to employ Hidden Markov Models (HMMs) to model them. Based on the HMM modeling, we investigate the relationship between model complexity and the generalization capability in downstream tasks. Specifically, we consider a popular tuning paradigm for downstream tasks, head tuning, where all pre-trained parameters are frozen and only individual heads are trained atop pre-trained LLMs. Our theoretical analysis reveals that the risk initially increases and then decreases with rising model complexity, showcasing a ``double descent'' phenomenon. In this case, the initial ``descent'' is degenerate, signifying that the ``sweet spot'' where bias and variance are balanced occurs when the model size is zero. Obtaining the presented in this study conclusion confronts several challenges, primarily revolving around effectively modeling autoregressive LLMs and downstream tasks, as well as conducting a comprehensive risk analysis for multivariate regression. Our research is substantiated by experiments conducted on data generated from HMMs, which provided empirical support and alignment with our theoretical insights.
\end{abstract}

%% file: sections/intro.tex
\section{Introduction}

Large Language Models (LLMs) have become a cornerstone in addressing a multitude of Natural Language Processing (NLP) tasks and have consistently demonstrated the state-of-the-art performance. Among the various approaches, the pre-trained fine-tuning paradigm stands out as the most prevalent and effective~\citep{NEURIPS2020_1457c0d6, DBLP:journals/corr/abs-1810-04805}, wherein fine-tuning involves training a fully connected network on a pre-trained LLM to tackle diverse downstream tasks. Recent research has underscored the significant impact of model complexity on the performance of fine-tuned LLMs~\citep{DBLP:journals/corr/abs-1912-02292, wei2023inverse}. However, the theoretical underpinnings of how model complexity influences fine-tuning performance remain unclear. In this paper, we delve into the intricate relationship between model complexity and the generalization capability of downstream tasks. We offer a theoretical analysis that elucidates the impact of increasing model size on the performance of LLMs. Our theoretical investigation aims to provide insights into the selection of an optimal model size to improve the performance of LLMs.

In practical applications, fine-tuning methods commonly involve adjusting the entire neural network~\citep{DBLP:journals/corr/abs-1810-04805}, and analyzing these methods theoretically can be challenging due to the intricate adjustments made to the parameters of pre-trained LLMs. However, a specific approach known as head tuning has garnered significant attention. Head tuning involves a scenario where all pre-trained parameters remain frozen, and only individual heads on top of the pre-trained LLMs are trained to enhance their performance on specific tasks. This methodology allows us to treat the pre-trained LLM as a black box~\citep{DBLP:journals/corr/abs-1802-05365}. Although head tuning simplifies the process, it still adheres to the fundamental pre-trained fine-tuning paradigm. Moreover, in situations involving substantial distribution shifts, head tuning has demonstrated the potential to outperform full fine-tuning~\citep{kumar2022finetuning}. Simultaneously, head tuning can be considered a form of representation learning, where we fine-tune a specific head on the representation of input tokens to address downstream tasks effectively. In summary, research on head tuning holds significant value and importance, attracting considerable interest from researchers~\citep{NEURIPS2021_86b3e165, DBLP:journals/corr/abs-2010-03648}.

While deep learning models like LLMs have showcased their formidable capabilities, our comprehension of their theoretical foundations remains somewhat limited. In previous studies exploring deep neural networks, there has been a tendency to simplify the analysis by considering single-layer linear settings~\citep{pmlr-v162-huang22e, DBLP:journals/corr/abs-2010-03648, NEURIPS2021_86b3e165, pmlr-v119-d-ascoli20a}. However, in the realm of deep learning, when we delve into what's known as the ``lazy regime"~\citep{NEURIPS2019_ae614c55}, where neural network weights remain close to their initial values during training, the deep neural network can be effectively linearized around its initial weights. Furthermore, it's important to note that under this regime, any transformer can be viewed as a Neural Tangent Kernel (NTK)~\citep{yang2020tensor}. An NTK is approximately equivalent to a random feature model~\citep{jacot2020neural}, which is linear in its parameters. These observations suggest that linear approximations can effectively capture predictive characteristics, even when dealing with distinct model structures. Simultaneously, it's important to note that although there have been numerous works investigating the relationship between model complexity and prediction risk under linear settings~\citep{doi:10.1137/20M1336072, 10.1214/21-AOS2133}, their applicability to our research is limited. This limitation stems from the unique characteristics of LLMs in our analysis, particularly our consideration of the model as an autoregressive one. In this context, we employ a Hidden Markov Model (HMM) to capture the autoregressive nature of LLMs, which sets our study apart from previous linear settings.

The HMM has a rich history in NLP and has found widespread applications, such as text tagging~\citep{merialdo-1994-tagging}, alignment~\citep{vogel1996hmm}, and language modeling~\citep{huang2011modeling, NEURIPS2021_86b3e165}. When dealing with the inherent complexity and irregularities of real-world data, it is essential to recognize the need for simplification. In the context of next word prediction tasks, each word within our corpus is probabilistically linked to every preceding word, reflecting the inherent dependencies in language. In this setup, it is common to consider the input data as the hidden states in an HMM, with the representations of the input data serving as the observed states in the HMM.  Considering the dependent structure in an HMM, our model can effectively capture essential representations of natural languages~\citep{10.1145/601858.601867, chiu-rush-2020-scaling}. Prior research has employed HMMs to investigate autoregressive models, demonstrating that head tuning can recover downstream labels~\citep{NEURIPS2021_86b3e165}. In contrast, our study focuses on establishing the next-word prediction risk and analyzing its relationship with model complexity. Additionally, there exist works that analyze regression errors in time series data without making the assumption of independence and identically distributed (i.i.d.) data generation processes~\citep{lee2023mean}. However, these studies do not incorporate the Hidden Markov assumption. Therefore, compared to their work, our research aligns more closely with NLP tasks, where the influence of sequential dependencies is crucial.

Specifically, our analysis assumes that we observe a sequence of tokens, which we employ to train an autoregressive LLM capable of predicting the next token. Subsequently, our downstream task revolves around learning a linear head to predict the next hidden state based on the predicted token. Our research commences with the establishment of our prediction risk, followed by a comprehensive bias-variance decomposition. To account for differences in the analysis, we divide our subsequent examination into two segments, focusing on the underparametrized case where the model size is smaller than the sample size and the overparametrized case where the model size is larger than the sample size, respectively. By combining insights from these two cases, we derive our estimate of the risk as a function of model complexity. Our findings reveal a phenomenon known as ``double descent"~\citep{Belkin_2019}, wherein the LLM's risk initially increases and then decreases with an increase in model complexity. Notably, the initial ``descent" is degenerate, indicating that the point at which bias and variance are balanced, the ``sweet spot" occurs when the model size is zero. Our primary contributions encompass the utilization of HMM to model autoregressive LLMs. We subsequently establish the next-word prediction risk and investigate its intricate relationship with model complexity. Additionally, we extend previous research on asymptotic risk in univariate regression to the realm of multiple regression. Under the assumptions we lay out, we provide an estimate for next-word prediction risk and offer practical suggestions for selecting an optimal model~size.

\subsection*{Related Work}
\paragraph{Large language models.}

In the realm of theoretical analysis of autoregressive LLMs, an extensive body of work has emerged. For instance, in~\citep{NEURIPS2021_02e656ad}, the authors established that, under their specified assumptions, training a linear head on autoregressive LLMs enables the prediction of one observed variable from another. Additionally, the work in~\citep{pmlr-v202-malladi23a} offers an explanation for the efficacy of parameter-efficient subspace-based fine-tuning methods for pre-trained language models, grounded in a kernel view. Furthermore, \citet{DBLP:journals/corr/abs-2010-03648} offers intuitive and mathematical explanations for the success of language model features in classification tasks by reinterpreting them as sentence completion problems.. Several other theoretical studies delve into the principles of self-supervised or contrastive learning, as evident in~\citep{pmlr-v97-saunshi19a, NEURIPS2021_27debb43, 10.5555/3546258.3546539, DBLP:journals/corr/abs-2010-03622, pmlr-v132-tosh21a}. Recent theoretical investigations have particularly focused on LLMs built upon the Transformer architecture, as exemplified by the prominent GPT series. For instance, in~\citep{feng2023revealing}, an analysis was conducted to explore how Chain-of-Thought prompting enhances the performance of Transformer-based LLMs from a computational perspective. Furthermore, the authors of~\citep{bai2023transformers} made significant strides in enhancing our understanding of the formidable in-context learning capabilities exhibited by Transformer models. These theoretical studies collectively contribute to a deeper comprehension of the principles underlying the robustness and efficacy of LLMs in various NLP tasks.

\paragraph{Double descent.}The double descent phenomenon was initially introduced by~\citep{Belkin_2019} and subsequently observed in various contexts, including~\citep{10.1214/21-AOS2133, ADVANI2020428, DBLP:journals/corr/abs-1810-09665, PhysRevE.100.012115}. Notably, researchers have documented instances of the double descent phenomenon in the context of model size in Transformer architectures as well~\citep{DBLP:journals/corr/abs-1912-02292}. To date, there has been a significant body of work focused on the theoretical analysis of double descent. Some of these analyses have been performed in the setting of linear least squares regression~\citep{doi:10.1137/20M1336072, 10.1214/21-AOS2133, doi:10.1073/pnas.1907378117, 9051968}. In the context of adversarial training, \citet{chen2020data} identified and demonstrated the existence of the double descent phenomenon using Gaussian and Bernoulli models. 
Under the same adversarial training setting, \citet{min2020curious} provided proof based on Gaussian models. 
Furthermore, \citet{javanmard2020precise} revealed the occurrence of double descent in adversarially robust linear regression. Additionally, the authors of~\citep{dar2020subspace} demonstrated that the generalization errors in corresponding subspace fitting problems follow double descent trends as the settings become more supervised and less orthonormally constrained. In the domain of reinforcement learning, \citet{NEURIPS2020_fdc42b6b} delved into the risk-sample tradeoff. However, the double descent phenomenon in deep learning still remains without a comprehensive explanation, as a theoretical proof for its occurrence has yet to be established by any researcher.

%% file: sections/formulation.tex
\section{Formulations and Notation}

In our investigation, the modeling of the next word prediction task holds significant importance. When confronted with the inherent complexity and irregularities of real-world data, it becomes crucial to simplify our approach. Specifically, within the realm of next word prediction tasks, each word in our corpus exhibits probabilistic associations with every preceding word, thus capturing the inherent linguistic dependencies. When a word is encountered, the probability distribution of the subsequent word is primarily influenced by the current word. To streamline our analysis, we employ a simplifying assumption in which the input word follows a Markov chain distribution. Within this simplified framework, the probability of the next word is exclusively dependent on the preceding word. By considering the dependent structure inherent in an HMM, our model can effectively capture essential representations of natural languages.~\citep{10.1145/601858.601867, chiu-rush-2020-scaling}.

\paragraph{Data distribution.}

Define $z_i^{1\times d}\in \{1,2,\dots,H\}^d$ as the $i$ th input token, which obeys the Markov chain distribution. First, we assume that $z_0\sim N(0, I_d)$, where $z_0$ is the first token of the input sentence. Define $A\in\mathbb{R}^{d\times d}$ the transition matrix of $z_i$. Subsequently, we establish continuous forms to represent the distribution of $z_i$.

\begin{align*}
    z_i=z_{i-1}A+\varepsilon_i, \quad \varepsilon_i\sim N(0,\sigma^2_\varepsilon I_d).
\end{align*}

It is important to highlight that in our context, $A$ represents the transition matrix. When $z_0$ is established, our objective is to introduce randomness into the distribution of subsequent tokens to better align with real-world scenarios. To achieve this, we employ a continuous approach by incorporating noise to enhance the level of randomness. It is noteworthy that $z_0$ follows a normal distribution with a mean of zero and an identity covariance matrix, denoted as $N(0, I_d)$. Subsequently, each $z_i$ is generated in accordance with a Markov Model. For any $0 \leq i \leq n$, $z_i$ follows a normal distribution with a mean of zero and a covariance matrix represented as $\Sigma_z$, where

\begin{align*}
    \Sigma_z&=\sum_{j=1}^{i}(A^T)^jA^j+I.
\end{align*}

\paragraph{Model setting.}
We assume that $x_i:=(x_{i1},...,x_{ip})$ is a representation of $z_i$, where $z_i$ is the $i$th input token. To simplify our analysis, we assume that it is a linear representation, which satisfy

\begin{align*}
    x_{i}=z_{i}W+u_{i},
\end{align*}

where $u_{i}:=(u_{i1},...,u_{ip})$ is a noisy, satisfying $u_{ij}\sim N(0, 1)$, and let $W\in\mathbb{R}^{d\times p} $ denote the representation matrix.  Define target $y_i^{1\times d}\in \{1,2,\dots,H\}^d$ to be the next word of $z_i$, which satisfy

\begin{align*}
    y_i=z_iA+\xi_i,
\end{align*}

where $\xi_i\in \mathbb{R}$ is also a noisy, satisfying $\xi_i\sim N(0,\sigma_{\xi}^2)$. Then we have $x_i\sim N(0, \Sigma_x)$ and $y_i\sim N(0, \Sigma_y)$, where

\begin{align*}
    \Sigma_y&=A^T(\sum_{j=1}^{i}(A^T)^jA^j+I)A+\Sigma_{\xi}, \\
    \Sigma_x&=W^T(\sum_{j=1}^{i}(A^T)^jA^j+I)W+I.
\end{align*}

When employing LLMs for next-word prediction, a common procedure involves feeding all word embeddings within a sentence into a pre-trained model to obtain their corresponding output representations. Subsequently, a feed-forward network is trained on the output representation of the last word, enabling the prediction of the subsequent word. In our framework, we assume the linearity of our pre-trained model, denoted as $W$, and we limit the feed-forward network to a single layer, denoted as $\widehat{B}\in\mathbb{R}^{p\times d}$. Given a sequence of tokens denoted as ${z_0, z_1, \dots, z_{n}}$, we derive their respective representations ${x_0, x_1, \dots, x_{n}}$. Our primary objective is to predict the next word, denoted as $y_{n}$, based on the representation $x_{n}$ through the training of a single-layer feed-forward network represented by $\widehat{B}$. 

\paragraph{Notation.}

We introduce the following notation for clarity in our discussion. We denote the dimension of the word embeddings as $H$. Consequently, we define $z_i^{1\times d}\in {1,2,\dots,H}^d$ as the representation of the $i$-th input token, $y_i^{1\times d}\in {1,2,\dots,H}^d$ as the subsequent word following $z_i$, and $x_i^{1\times p}$ as the representation of $z_i$. Specifically, $d$ signifies the dimensionality of vectors $z$ and $y$, while $p$ represents the dimensionality of vector $x$. We denote the sample size of $x$ as $n$ and define the overparametrization ratio $\gamma=p/n$. Let $\Sigma_\alpha$ represent the covariance matrix associated with variable $\alpha$, where $\alpha$ can take on values such as $x$, $y$, $z$, $\epsilon$, and $\xi$. Furthermore, we introduce $M^+$ to denote the Moore-Penrose inverse of the matrix $M$. We make the assumption that the input vector $z$ adheres to a Markov chain distribution, and we introduce $A\in\mathbb{R}^{d\times d}$ to denote the transition matrix. Additionally, we employ the symbols $W$ to represent the representation matrix and $\widehat{B}$ to signify a single-layer feed-forward network. We introduce the concepts of bias and variance within the context of next-word prediction risk and ridge regression risk. Specifically, we define $B_X(\widehat{B};B)$ to represent the bias in next-word prediction risk, while $V_X(\widehat{B};B)$ quantifies the variance. Furthermore, we provide asymptotic approximations for these measures denoted as $\mathscr{B}(\gamma,\Sigma_x)$ and $\mathscr{V}(\gamma,\Sigma_x)$. Expanding our discussion to encompass the ridge regression scenario, we introduce $B_X(\widehat{B}_{\lambda};B)$ as the bias associated with ridge regression risk and $V_X(\widehat{B}_{\lambda};B)$ as the corresponding variance. In this context, we also present asymptotic approximations, which are expressed as $\mathscr{B}(\lambda;\gamma, \Sigma_x)$ and $\mathscr{V}(\lambda;\gamma, \Sigma_x)$. 

\paragraph{Remark.}We want to emphasize that our research should not be viewed as a mere extension of the previous work~\citep{10.1214/21-AOS2133}. Our approach differs significantly, as we assume that the data originates from a HMM, which moves beyond the basic i.i.d.~framework used in the previous work. Furthermore, our analysis focuses on next-word prediction, which represents a progression from a univariate to a multivariate regression model. These aspects introduce significant challenges, particularly in analyzing asymptotic bias and variance, thereby distinguishing our work from the previous study~\citep{10.1214/21-AOS2133}. 

Moreover, our findings, which rely on linear models, have the potential to be extended to cases with a multi-layer head, such as transformers. This extension is viable because any transformer can be regarded as a NTK under the lazy regime~\citep{yang2020tensor}, and an NTK is approximately equivalent to a random feature model~\citep{jacot2020neural}, which is linear in its parameters. Consequently, the use of linear approximation from embeddings to outputs is not an oversimplification but rather a widely recognized approach in the field, as evidenced by other studies\citep{NEURIPS2021_86b3e165}.

%% file: sections/analysis.tex
\section{Analysis}

In this section, we embark on an investigation into the risk associated with next-word prediction. Our analytical exploration comprises three key components. In the first part, we lay the foundational groundwork. Through a covariance alignment between the representation and the subsequent word, we refine our model. Following this, we introduce the concept of next-word prediction risk and provide a comprehensive bias-variance decomposition. 

Notably, the behavior of the matrix $X^TX$ exhibits an invertible nature when the number of model size $p$ is less than the sample size $n$, and it becomes singular when $p$ exceeds $n$. Consequently, our analysis diverges into two distinct segments. In the second part, we delve into the scenario of underparametrization, where $p<n$. In the third part, we explore the overparametrized case, where $p>n$. Finally, we synthesize the insights gained from these two cases to formulate our ultimate conclusion.

\subsection{Foundational Preparation}

Recall the distribution of the $i$ th input token $z_i$, $z_i$'s next word $y_i$ and $z_i$'s representation $x_i$. Regress $y_i$ on $x_i$, we can rewrite our model as 

\begin{align*}
    y_i=x_iB+\epsilon, 
\end{align*}

where

\begin{align*}
    B&=W^T(\sum_{j=1}^{i}(A^T)^jA^j+I)(I+WW^T(\sum_{j=1}^{i}(A^T)^jA^j+I))^{-1}A,\quad \epsilon\sim N(0, \Sigma_{\epsilon}), \\ 
    \Sigma_{\epsilon}&=\Sigma_{\xi}+A^T(I+(\sum_{j=1}^{i}(A^T)^jA^j+I)WW^T)^{-1}(\sum_{j=1}^{i}(A^T)^jA^j+I)A.
\end{align*}

In this step, we adopt an approach akin to that of~\citep{10.1214/21-AOS2133} by aligning the covariances between these variables. In this paper, we focus on training a single-layer feed-forward network on the output of pre-trained LLMs, which corresponds to the representation $x_i$. Consequently, we assume that the training process is effective, resulting in a training error of zero. Thus, we can regard our single-layer feed-forward network after training as a least squares estimator $\widehat{B}$ with $n$ training cases, where:

\begin{align*}
    \widehat{B}=(X^TX)^+X^TY.
\end{align*}

Here $X^{n\times p}$ is the representation matrix with rows $x_i$, and Y is the target matrix with rows $y_i$, where $0 \leq i \leq n-1$. The notation $(X^TX)^+$ refers to the Moore-Penrose inverse of the matrix $X^TX$. Consider a test sample $(x_{n}, y_{n})$, for an estimator $\widehat{B}$, we define the out-of-sample next word prediction risk as

\begin{align*}
    R_X(\widehat{B};B)=E({\rm Tr}[(x_{n}B-x_{n}\widehat{B})^T(x_{n}B-x_{n}\widehat{B})]).
\end{align*}

Also we have the bias-variance decomposition

\begin{align*}
    R_X(\widehat{B};B)&=E({\rm Tr}((x_{n}B-E(x_{n}\widehat{B}))^T(x_{n}B-E(x_{n}\widehat{B}))))\\
    &+E({\rm Tr}((E(x_{n}\widehat{B})-x_{n}\widehat{B})^T(E(x_{n}\widehat{B})-x_{n}\widehat{B}))).
\end{align*}

The conventional bias-variance decomposition historically applied to univariate regression. In contrast, our approach involves multiple regression, signifying a departure from past practices. Consequently, here we have extended the traditional bias-variance decomposition, originally designed for single regression, to encompass the realm of multiple regression.

\subsection{Underparametrized Asymptotics}

In this subsection we first consider the underparametrized case, which means $p<n$. We consider an asymptotic setup that $n, p\rightarrow\infty $, and in such a way that $p/n\rightarrow\gamma$. To establish the subsequent theorem, we first present the following lemma, which is straightforward to demonstrate. 

\begin{lemma}\label{lem: 1}
    For the least squares estimator $\widehat{B}$, the next word prediction risk has bias defined as $B_X(\widehat{B};B)$, and variance defined as $V_X(\widehat{B};B)$ .
    
\begin{align*}
    B_X(\widehat{B};B)&={\rm Tr}(B^T(X^TX(X^TX)^+-I)\Sigma_x((X^TX)^+X^TX-I)B),\\
    V_X(\widehat{B};B)&={\rm Tr}(\Sigma_{\epsilon}){\rm Tr}(\Sigma_x (X^TX)^+).
\end{align*}

\end{lemma}

\textit{Proof.} Notice that $E(\widehat{B})=(X^TX)^+X^TXB$, and recall that $\widehat{B}=(X^TX)^+X^TY$, hence, it is straightforward to derive this lemma.

Under the underparameterized assumption, it is worth noting that $X^TX$ is invertible, and as evident from the aforementioned lemma, the bias effectively amounts to zero in this scenario. Given this observation and harking back to the bias-variance decomposition of the next word prediction risk, we can now state the following theorem.

\begin{theorem}\label{thm: 1}
    Assume that $\lambda_{min}(\Sigma_x)\geq c > 0$ for all n, p and a constant c, then as $n, p\rightarrow\infty $, in such a way that $p/n\rightarrow\gamma < 1$ , the next word prediction risk satisfies, almost surely

\begin{align*}
    \lim_{n\rightarrow\infty}R_X(\widehat{B};B)={\rm Tr}(\Sigma_{\epsilon})\frac{\gamma}{1-\gamma}.
\end{align*}

\end{theorem}
We will complete the proof in Appendix \ref{sec: A}.

As an initial observation, it is important to highlight that in the underparametrized regime, where the ratio $\gamma<1$, the minimum-norm estimator aligns with the conventional least squares estimator. In this scenario, the primary source of risk is attributed solely to variance, and there is no bias. Moreover, it is noteworthy that this risk does not exhibit any dependency on the parameters $B$ and $\Sigma_x$. Interestingly, as we approach the interpolation boundary, where $\gamma$ tends towards 1, the asymptotic risk experiences a significant increase.

It is worth noting that, as depicted in Figure \ref{figure 1}, the risk curve of Theorem \ref{thm: 1} in the underparameterized regime does not conform to the typical U-shaped pattern. It is essential to acknowledge that the U-shaped curve does not universally manifest, as discussed by~\citep{NEURIPS2021_4ae67a7d}. For instance, there is no U-shaped curve evident in Figure 1 of~\citep{doi:10.1137/20M1336072}. In practical applications,  such as when employing Transformers for language translation (e.g., IWSLT’14 German-to-English) as documented in~\citep{DBLP:journals/corr/abs-1912-02292}, the classical U-shaped curve is notably absent. Simultaneously, the absence of the U-shaped curve was also observed in our own simulation experiments.

\subsection{Overparametrized Asymptotics}

In this subsection, we delve into the overparametrized scenario, where the presence of non-zero bias is notable. The next-word prediction risk is intricately tied to the geometric properties of $\Sigma_x$. To elucidate, we introduce the eigenvalues of $\Sigma_x$ as ${s_1, s_2, \dots, s_p}$, where $s_1\geq s_2\geq\dots\geq s_p\geq0$. Subsequently, we articulate our assumptions concerning our model, where $M$ represents a substantial constant.

\begin{assumption}

Recall the eigenvalue of $\Sigma_x$ and the overparametrization ratio $\gamma=p/n$ , we have 

(1) $s_1\leq M$, $\sum_{i=1}^{p}\frac{1}{s_i}\leq M$.

(2) $|1-p/n|\geq 1/M$, $1/M\leq p/n\leq M$.

\end{assumption}

 Assumption(1) necessitate that the eigenvalues of $\Sigma_x$ exhibit bounded characteristics, while also avoiding accumulation in close proximity to zero.And assumption(2) mandates that the ratio of $p$ to $n$ remains within bounds, while also maintaining a separation from the critical threshold of interpolation, denoted by $p/n=1$.

\begin{definition}\label{def: 1}
    Define $\gamma=p/n$, for $\gamma\in \mathbb{R}_{>1}$, define $c_0$ to be the unique non-negative solution to

\begin{align}\label{equ: 1}
    1-\frac{1}{\gamma}=\sum_{i=1}^{p}\frac{1}{1+c_0\gamma s_i}.
\end{align}

\end{definition}

When the value of $\gamma$ exceeds one, the left-hand side of the equation mentioned above remains a constant within the range of zero to one. 
On the other hand, the function on the right-hand side displays monotonic behavior concerning the parameter $c_0$. As $c_0$ approaches zero, the function value tends toward one, and as $c_0$ approaches infinity, the function value tends toward zero. This characteristic implies that the equation must possess a unique non-negative solution.

However, it's important to note that when the eigenvalue composition of $\Sigma_x$ becomes complex, solving the analytical solution of the equation can become significantly challenging. In such cases, the solution may require advanced numerical methods or specialized techniques to obtain accurate results. And then we define the bias and variance of the next word prediction risk.

\begin{definition}\label{def: 2}
We define bias as $\mathscr{B}(\gamma,\Sigma_x)$ and variance as $\mathscr{V}(\gamma,\Sigma_x)$, as shown below

\begin{align*}
    \mathscr{B}(\gamma,\Sigma_x)&:=\{1+\gamma c_0\frac{\sum_{i=1}^{p} \frac{s_i^2}{(1+c_0\gamma s_i)^2}}{\sum_{i=1}^{p}\frac{s_i}{(1+c_0\gamma s_i)^2}}\}{\rm Tr}(B^T(I+c_0\gamma\Sigma_x)^{-2}\Sigma_x B),\\
    \mathscr{V}(\gamma,\Sigma_x)&:={\rm Tr}(\Sigma_{\epsilon})\gamma c_0\frac{\sum_{i=1}^{p} \frac{s_i^2}{(1+c_0\gamma s_i)^2}}{\sum_{i=1}^{p}\frac{s_i}{(1+c_0\gamma s_i)^2}}.
\end{align*}

\end{definition}

It is worth noting that the most intricate aspect of numerically evaluating $\mathscr{B}(\gamma,\Sigma_x)$ and $\mathscr{V}(\gamma,\Sigma_x)$ lies in determining the unique non-negative solution of \eqref{equ: 1}. The subsequent theorem establishes that, under appropriate technical conditions, the functions $\mathscr{B}(\gamma,\Sigma_x)$ and $\mathscr{V}(\gamma,\Sigma_x)$ play a pivotal role in characterizing the test error. It's worth noting that similar theorems in the context of univariate regression have been previously established in prior research~\citep{10.1214/21-AOS2133}. We extend these findings to the domain of multiple regression.

\begin{theorem}\label{thm: 2}
Assume that $\lambda_{min}(\Sigma_x)>1/M$, define $\gamma=p/n$, then for any constants $D>0$ there exist $C=C(M, D)$, with probability at least $1-Cn^{-D}$ the following hold

\begin{align*} 
    &R_X(\widehat{B};B)=B_X(\widehat{B};B)+V_X(\widehat{B};B),\\
    &|B_X(\widehat{B};B)-\mathscr{B}(\gamma,\Sigma_x)|\leq C\frac{(||B||_F+1)}{n^{1/4}}, \\
    &|V_X(\widehat{B};B)- \mathscr{V}(\gamma,\Sigma_x)|\leq C(\frac{1}{n^{1/4}}+1).
\end{align*}

\end{theorem}

We prove this theorem in Appendix \ref{sec: C}.

Theorem~\ref{thm: 2} establishes deterministic approximations for both bias and variance, which remain valid even for finite values of $n$ and $p$, with $\gamma=p/n$ being a non-asymptotic quantity. Moreover, it's noteworthy that the error bounds demonstrated here display uniformity, signifying a pronounced dependence on the constant $M$. This characteristic sets it apart from the asymptotic context found in the works of~\citep{richards2021asymptotics, NEURIPS2020_72e6d323}.

\begin{figure}[t]
\centering
\includegraphics[width=0.5\linewidth]{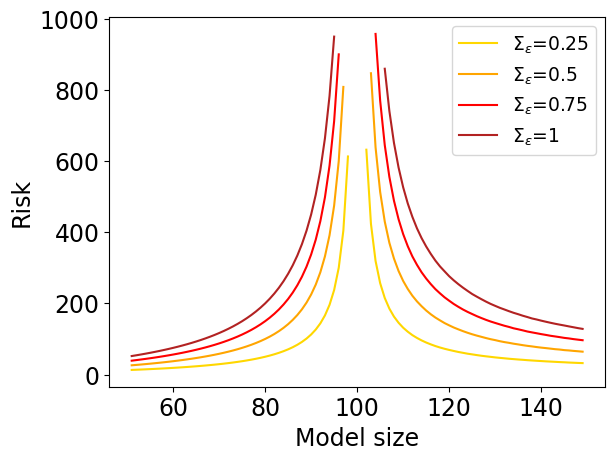}
\caption{The risk curves illustrate the behavior of the least squares estimator $\widehat{B}$ concerning its dependence on the model size, denoted as $p$, which is also the dimensionality of the vector $x$. In this context, $p$ varies within the range of 50 to 150. These curves are based on a scenario with $n=100$, $d=50$, and varying levels of noise variance $\Sigma_{\epsilon}$. The continuous lines represent the analytical predictions derived from Definition~\ref{def: 2}.}
\label{figure 1}
\end{figure}

To establish the proof of Theorem \ref{thm: 2}, we take an intermediate step by proving the following theorem within the context of ridge regression. In the case of ridge regression, under the same settings as those in the least squares estimator mentioned earlier, we introduce the ridge regression estimator denoted as $\widehat{B}_{\lambda}$. This estimator is computed as $\widehat{B}_{\lambda} = (X^TX + n\lambda I)^{-1}X^TY$. It is essential to highlight that even though we consider our trained single-layer feed-forward network as a least squares estimator $\widehat{B}$, it can also be referred to as the ``ridgeless" least squares estimator, due to the fact that $\widehat{B} = \lim_{\lambda\rightarrow 0^{+}} \widehat{B}_{\lambda}$. After completing the proof of Theorem \ref{thm: 3}, we will utilize this property of $\widehat{B}$ to establish the proof of Theorem \ref{thm: 2}. 

To facilitate our proof, we provide the subsequent definitions of bias and variance within the ridge regression framework.

\begin{definition}\label{def: 3}
    For $\gamma\in \mathbb{R}_{>1}$ and $z \in \mathbb{C_+}$ , where $\mathbb{C_+}$ represents the set of complex numbers with $Im(z) > 0$, we define $m_n(z)$ as the unique solution to the following equation
    
\begin{align}\label{equ: 2}
    m_n(z)=\sum_{i=1}^{p}\frac{1}{[1-\gamma-\gamma zm_n(z)]s_i-z}.
\end{align}

Additionally, we define $m_{n,1}(z)$ as follows

\begin{align}\label{equ: 3}
    m_{n,1}(z):=\frac{\sum_{i=1}^{p}\frac{[1-\gamma-\gamma zm_n(z)]s_i^2}{[[1-\gamma-\gamma zm_n(z)]s_i-z]^2}}{1-\gamma\sum_{i=1}^{p}\frac{zs_i}{[[1-\gamma-\gamma zm_n(z)]s_i-z]^2}}.
\end{align}

We proceed to define the prediction bias of ridge regression as $\mathscr{B}(\lambda;\gamma, \Sigma_x)$ and the variance as $\mathscr{V}(\lambda;\gamma, \Sigma_x)$ through the following expressions

\begin{align*}
    \mathscr{B}(\lambda;\gamma, \Sigma_x)&:=\lambda^2(1+\gamma m_{n, 1}(-\lambda)){\rm Tr}(B^T(\lambda I+(1-\gamma+\gamma\lambda m_n(-\lambda))\Sigma_x)^{-2}\Sigma_x B),\\
    \mathscr{V}(\lambda;\gamma, \Sigma_x)&:={\rm Tr}(\Sigma_{\epsilon})\gamma \sum_{i=1}^{p}\frac{s_i^2(1-\gamma+\gamma\lambda^2 m_n^{'}(-\lambda))}{[\lambda+s_i(1-\gamma+\gamma\lambda m_n(-\lambda))]^2}.
\end{align*}

\end{definition}

Next, we present our deterministic approximation of the prediction risk associated with ridge regression.

\begin{theorem}\label{thm: 3}
    Assuming that Assumption 1 is satisfied, and given that $\lambda_{min}(\Sigma_x)>1/M$, we define $\gamma=p/n$. For any constants $D>0$ (which can be arbitrarily large) and any $\epsilon>0$ (which can be arbitrarily small), there exists a constant $C=C(M, D)$ such that, with a probability of at least $1-Cn^{-D}$, the following conditions are met:

\begin{align*}
    &R_X(\widehat{B}_{\lambda};B)=B_X(\widehat{B}_{\lambda};B)+V_X(\widehat{B}_{\lambda};B), \\
    &|B_X(\widehat{B}_{\lambda};B)-\mathscr{B}(\lambda;\gamma, \Sigma_x)|\leq\frac{C||B||_F}{n^{(1-\epsilon)/2}} \lambda, \\
    &|V_X(\widehat{B}_{\lambda};B)-\mathscr{V}(\lambda;\gamma, \Sigma_x)|\leq\frac{C}{\lambda^2n^{(1-\epsilon)/2}}.
\end{align*}

\end{theorem}

The proof of this theorem is deferred to Appendix \ref{sec: B}.

Specifically, Theorem \ref{thm: 3} provides non-asymptotic deterministic approximations for the ridge regression prediction bias $B_X(\widehat{B}_{\lambda};B)$ and variance $V_X(\widehat{B}_{\lambda};B)$. Notably, the error terms in these approximations exhibit uniformity with respect to the covariance matrix $\Sigma_x$ and display a near-optimal dependence on the sample size $n$. Having established Theorem \ref{thm: 3}, we can now utilize it as a stepping stone to demonstrate the preceding Theorem \ref{thm: 2}.

\begin{figure}[t]
\centering
\includegraphics[width=0.5\linewidth]{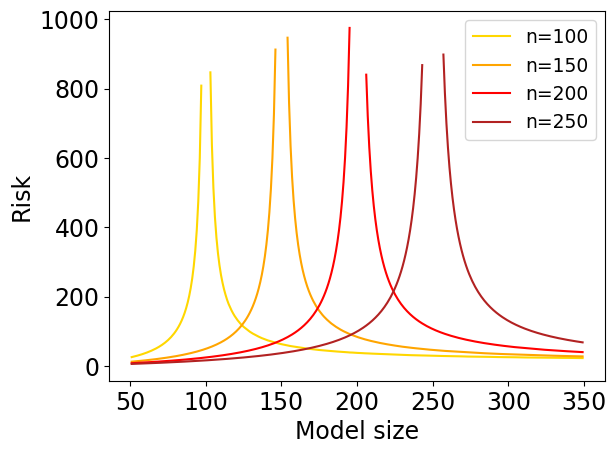}
\caption{The risk curves illustrate the behavior of the least squares estimator $\widehat{B}$ concerning its dependence on the model size, denoted as $p$, which is also the dimensionality of the vector $x$. In this context, $p$ varies within the range of 50 to 350. These curves are based on a scenario with $\Sigma_{\epsilon}=0.5$, $d=50$, and varying levels of data samples $n$. The continuous lines represent the analytical predictions derived from Definition \ref{def: 2}.}
\label{figure 2}
\end{figure}

\paragraph{Remark.}

By combining the insights from Theorem \ref{thm: 1} and Theorem \ref{thm: 2}, we arrive at our estimation of the next-word prediction risk, as illustrated in Figure \ref{figure 1}. It becomes evident that as the model size $p$ increases, the risk exhibits a notable increase, particularly in proximity to the interpolation threshold $p/n = 1$. However, when the model size surpasses the number of input samples ($p > n$), we observe a subsequent decrease in risk. Notably, when the volume of input samples denoted by $n$ expands, there is a corresponding requirement for the model size $p$ to increase in order to reach the interpolation threshold. This phenomenon leads to a displacement of the threshold point towards the right, as visualized in Figure \ref{figure 2}. These findings elucidate the underlying reasons for the occurrence of double descent behavior during the task of next-word prediction by LLMs and elucidate how augmenting the number of samples leads to a rightward shift in the peak of the test error, as previously discussed in~\citet{DBLP:journals/corr/abs-1912-02292}. 

In fact, our theory could still hold when the HMM assumption is relaxed. For instance, one can relax the Markov assumption in HMMs, where $z_i=z_{i-1}A+\varepsilon$, into the case of more extended dependencies, such as $z_i=(z_{i-1},\dots,z_{i-k})A+\varepsilon$, where $k$ is a fixed positive integer. Upon a careful review of our proof, we have found that our main results remain essentially the same with this adjustment. The primary difference is an increase in the number of columns, now adjusted to $kd$. We will consider other popular extensions of HMMs, such as auto-regressive HMM capturing long-term dependencies~\citep{pmlr-v48-guan16}, as future work.

In light of our theoretical findings, we offer practical guidance on optimizing the trade-off between model size and training data volume. Our theoretical analysis indicates that the prediction error is minimized when the parameter count $p$ is smaller than the number of training data samples $n$. Consequently, we advocate that for enhanced performance when employing LLMs to tackle downstream NLP tasks, the volume of training data should significantly surpass the model size. Indeed, this recommendation aligns with the empirical findings presented in~\citet{hoffmann2022training} and is reflective of the emerging trend in the utilization of LLMs, as evidenced by recent work in the field~\citep{touvron2023llama}.

%% file: sections/simulation.tex
\section{Simulation}

\begin{figure}[t]
\centering
\includegraphics[width=0.5\linewidth]{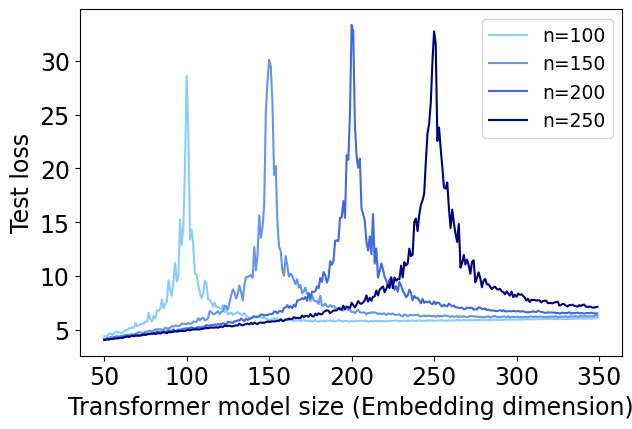}
\caption{The cross-entropy test loss curve is generated by employing prediction through head tuning on the Transformer model. In this experiment, both the training and test data are generated using HMM.}
\label{figure 3}
\end{figure}

In the preceding section, we established that the implementation of head tuning on pre-trained LLMs leads to a double descent phenomenon with the initial ``descent" is degenerate, with the initial ``descent" being degenerate. In the subsequent section, we aim to empirically validate our theoretical findings by applying head tuning to an encoder-decoder Transformer model using data generated by a HMM. Our objective is to corroborate our theoretical conclusions within a more practical and intricate context, where the representation of input tokens is acquired through a Transformer.

\paragraph{Training data and downstream task.}In our experimental setup, we make an initial assumption that the first token, denoted as $z_0$, follows a standard normal distribution with a mean of zero and an identity covariance matrix, represented as $N(0, I_d)$. Subsequently, we employ a HMM for generating the succeeding tokens. The transition matrix of the HMM is initialized randomly. Notably, we restrict our input to a single token at a time, and the primary objective of the downstream task is to predict the subsequent token based on the provided input token. Throughout our experiments, we will adjust the size of the hidden state within the HMM in the range of 50 to 150, in alignment with changes in the hidden dimension of the Transformer model.

\paragraph{Model training.}Our approach to tackling downstream tasks involves conducting head tuning using an encoder-decoder Transformer architecture. In our experimental setup, we employ Transformer models with specific configurations: $model \ depths = 1$, $heads = 1$, and hidden size varies within the range of 50 to 350. Initially, we initialize the parameters of the Transformer randomly and maintain them as fixed, treating the Transformer as a pre-trained model. Subsequently, we train a single-layer feed-forward network using the representations generated by the Transformer to predict the subsequent token. To investigate the influence of model size, we systematically augment the Transformer's capacity by increasing its hidden size. This augmentation is carried out incrementally, alongside corresponding adjustments to the hidden state size of the HMM. Notably, for each unique combination of hidden size and hidden state size, we conduct separate training. Each test dataset includes $n_{test}=20$ samples. To better support our theoretical conclusions, we selected four different numbers of training samples $n_{train}$: 100, 150, 200, and 250. We evaluate the model's performance using the cross-entropy loss as the metric.

\paragraph{Results.}Figure \ref{figure 3} illustrates our experimental findings, which reveal a notable trend: as the size of the Transformer model increases, the test error initially rises before subsequently declining. This observed pattern closely aligns with the conclusions drawn from our theoretical analysis. Additionally, it is noteworthy that the test loss is minimized when the Transformer model size is smaller than the number of training data samples, which further aligns with the conclusions drawn from our theoretical analysis.

%% file: sections/conclusion.tex
\section{Conclusion}

In conclusion, our study delves into the intricate relationship between model complexity and fine-tuning performance in autoregressive LLMs. By employing HMMs to analyze autoregressive LLMs, we uncover a striking ``double descent'' pattern. This finding highlights the intricate interplay of model intricacy and generalization capacity in downstream tasks, particularly in the context of head tuning. 

Nonetheless, our study comes with certain limitations. We restricted our investigation to a linear setting, which is a simplified approximation and may not fully capture the complexities of actual LLMs. Additionally, our analysis did not encompass prompt tuning, a highly effective fine-tuning technique that has garnered significant attention within the field. Extending our findings to address these limitations would be a valuable direction for future research.

%% file: sections/appendix.tex
\appendix
\begin{center}
    \bf\Large Appendix
\end{center}

\section{Proof of Theorem 1}\label{sec: A}
Based on Lemma \ref{lem: 1}, it is evident that
\begin{align*}
    B_X(\widehat{B};B)&={\rm Tr}(B^T(X^TX(X^TX)^+-I)\Sigma((X^TX)^+X^TX-I)B)\\
    &=0.
\end{align*}
Simultaneously, it becomes apparent that
\begin{align*}
    V_X(\widehat{B};B)&={\rm Tr}(\Sigma_{\epsilon}){\rm Tr}(\Sigma (X^TX)^+X^TX(X^TX)^+)\\
    &={\rm Tr}(\Sigma_{\epsilon}){\rm Tr}(\Sigma (X^TX)^+)\\
    &\rightarrow{\rm Tr}(\Sigma_{\epsilon})\frac{\gamma}{1-\gamma}.
\end{align*}
The final step can be derived from the proof of Proposition 2 presented in~\citep{10.1214/21-AOS2133}. In summary, we can obtain
\begin{align*}
    \lim_{n\rightarrow\infty}R_X(\widehat{B};B)={\rm Tr}(\Sigma_{\epsilon})\frac{\gamma}{1-\gamma}.
\end{align*}

\section{Proof of Theorem 3}\label{sec: B}
As the proof of Theorem \ref{def: 2} relies on the foundation laid by Theorem \ref{thm: 3}, we will commence by presenting the proof of Theorem \ref{thm: 3}.

\subsection{Proof of Theorem 3: Bias term}\label{sec: B.1}
We will begin by providing a lemma that will serve as a useful tool for subsequent proofs.\\

\begin{lemma}\label{lem: 2}
    For any square matrix $A^{p\times p}$ and matrix $B^{p\times d}$, there exists a vector $\beta^{p\times1}$ such that $||\beta||_2=||B||_F$, and this vector satisfies:
\begin{align*}
    {\rm Tr}(B^TAB)\leq \beta^TA\beta.
\end{align*}
\end{lemma}

\textit{Proof.} Let
\begin{align*}
    B=\begin{pmatrix}
        b_{11}&b_{12}&\cdots&b_{1d} \\
        b_{21}&b_{22}&\cdots&b_{2d} \\
        \vdots&\vdots&\ddots&\vdots \\
        b_{p1}&b_{p2}&\cdots&b_{pd}
      \end{pmatrix}, 
\end{align*}

\begin{align*}
    \quad \beta = (\beta_1, \beta_2, \cdots, \beta_p)^T,
\end{align*}
and further let
\begin{align*}
    A=\begin{pmatrix}
        a_{11}&a_{12}&\cdots&a_{1p} \\
        a_{21}&a_{22}&\cdots&a_{2p} \\
        \vdots&\vdots&\ddots&\vdots \\
        a_{p1}&a_{p2}&\cdots&a_{pp}
      \end{pmatrix}.
\end{align*}
Consequently, we can derive that
\begin{align*}
    \beta^TA\beta = \sum_{i = 1}^{p}\sum_{j=1}^{p}a_{ij}\beta_i\beta_j.
\end{align*}
Moreover, we can also establish that
\begin{align*}
    {\rm Tr}(B^TAB)&={\rm Tr}(ABB^T) \\
    &={\rm Tr}(A\begin{pmatrix}
        \sum_{j=1}^{d}b_{1j}^2&\sum_{j=1}^{d}b_{1j}b_{2j}&\cdots&\sum_{j=1}^{d}b_{1j}b_{pj} \\
        \sum_{j=1}^{d}b_{2j}b_{1j}&\sum_{j=1}^{d}b_{2j}^2&\cdots&\sum_{j=1}^{d}b_{2j}b_{pj} \\
        \vdots&\vdots&\ddots&\vdots \\
        \sum_{j=1}^{d}b_{pj}b_{1j}&\sum_{j=1}^{d}b_{pj}b_{2j}&\cdots&\sum_{j=1}^{d}b_{pj}^2
      \end{pmatrix}) \\
    &=\sum_{i = 1}^{p}\sum_{j=1}^{p}a_{ij}\sum_{m=1}^{d}b_{jm}b_{im}.
\end{align*}

If we assume that $\beta_i^2=\sum_{j=1}^{d}b_{ij}$ for $i=1, 2, \cdots, p$, then it follows that $||\beta||_2=||B||_F$. Subsequently, we can readily derive that

\begin{align*}
    \sum_{m=1}^{d}b_{jm}b_{im}\leq \beta_i\beta_j, \quad 1\leq i, j \leq p.
\end{align*}

Hence, we can conclude that ${\rm Tr}(B^TAB)\leq \beta^TA\beta$.\\

We will now commence with the proof of the bias term in Theorem \ref{thm: 3}. To begin, let's revisit the expressions for bias within our ridge regression framework at the regularization parameter $\lambda$:
\begin{align*}
    B_X(\widehat{B}_{\lambda};B)=\lambda^2{\rm Tr}(B^T(S_X+\lambda I)^{-1}\Sigma(S_X+\lambda I)^{-1}B).
\end{align*}
Then we define
\begin{align*}
    S_Z:=\frac{Z^TZ}{n} &, \quad S_X:=\frac{X^TX}{n}=\Sigma^{1/2} S_Z \Sigma^{1/2}. \\
\end{align*}
Building upon these expressions, we proceed to define
\begin{align*}
    \overline{F}_n(\eta, \lambda)&:={\rm Tr}(\lambda B^T(S_X+\lambda I+\lambda\eta\Sigma)^{-1}B) \\
    &={\rm Tr}(\lambda B_\eta^T(\Sigma_\eta^{1/2} S_Z \Sigma_\eta^{1/2}+\lambda I)^{-1}B_\eta), 
\end{align*}
where
\begin{align*}
    \Sigma_\eta:=\Sigma(I+\eta\Sigma)^{-1}, \quad B_\eta:=(I+\eta\Sigma)^{-1/2}B.
\end{align*}
Then we can get
\begin{align*}
    -\frac{\partial \overline{F}_n}{\partial \eta}(0, \lambda)=B_X(\widehat{B}_{\lambda};B).
\end{align*}
We define
\begin{align*}
    {F}_n(\eta, \lambda):=-{\rm Tr}(B_\eta^T(I+r_n(-\lambda,\eta)\Sigma\eta)^{-1}B_\eta),
\end{align*}
where $r_n=r_n(-\lambda,\eta)$ represents the solution to the following equation:
\begin{align*}
    \frac{1}{r_n}=\lambda+\gamma \frac{1}{p}\sum_{i=1}^{p}\frac{s_i(\eta)}{1+s_i(\eta)r_n}.
\end{align*}
Recalling the expression of the prediction bias as defined in Definition \ref{def: 3}, it becomes evident that
\begin{align*}
    -\frac{\partial F_n}{\partial \eta}(0, \lambda)=\mathscr{B}_X(\widehat{B}_{\lambda};B).
\end{align*}
We introduce the notation $s_1(\eta)\geq s_2(\eta)\geq s_3(\eta)\geq\cdots\geq s_p(\eta)$ to represent the eigenvalues of $\Sigma_\eta$. Then, we define $r_n=r_n(z, \eta)$ as the unique solution of:
\begin{align*}
    \frac{1}{r_n}=-z+\gamma\frac{1}{p}\sum_{i=1}^p\frac{s_i(\eta)}{1+s_i(\eta)}.
\end{align*}
By employing Lemma \ref{lem: 2}, It is straightforward to establish that there exists a vector $\beta$, where $||\beta||_2=||B||_F$, and this $\beta$ satisfies
\begin{align*}
    |\frac{\partial\overline{F}_n}{\partial\eta}(0, \lambda)-\frac{\partial F_n}{\partial\eta}(0, \lambda)| &= |\lambda^2{\rm Tr}(B^T(S_X+\lambda I)^{-1}\Sigma(S_X+\lambda I)^{-1}B) \\
    &-\lambda^2(1+\gamma m_{n, 1}(-\lambda)){\rm Tr}(B^T(\lambda I+(1-\gamma+\gamma\lambda m_n(-\lambda))\Sigma)^{-2}\Sigma B) |\\
    &\leq |\lambda^2\beta^T(S_X+\lambda I)^{-1}\Sigma(S_X+\lambda I)^{-1}\beta \\
    &-\lambda^2(1+\gamma m_{n, 1}(-\lambda))\beta^T(\lambda I+(1-\gamma+\gamma\lambda m_n(-\lambda))\Sigma)^{-2}\Sigma \beta| \\
    &= |\frac{\partial(\lambda \beta_\eta^T(\Sigma_\eta^{1/2} S_Z \Sigma_\eta^{1/2}+\lambda I)^{-1}\beta_\eta)}{\partial\eta}|_{\eta=0}-\frac{\partial(-\beta_\eta^T(I+r_n(-\lambda,\eta)\Sigma\eta)^{-1}\beta_\eta)}{\partial}|_{\eta=0}| \\
    &\leq \frac{C||B||_F}{n^{(1-\epsilon)/2}\lambda}, \quad \forall\lambda\in (n^{-2/3+\epsilon_0}, \infty),  \eta\in(-\frac{1}{2M}, \infty).
\end{align*}
The final step in this process can be derived from equation (85) in~\citep{10.1214/21-AOS2133}. 
Note that
\begin{align*}
    |B_X(\widehat{B}_{\lambda};B)-\mathscr{B}(\lambda;\gamma, \Sigma_x)|=|\frac{\partial\overline{F}_n}{\partial\eta}(0, \lambda)-\frac{\partial F_n}{\partial\eta}(0, \lambda)|,
\end{align*}
thus, we have completed the proof of the bias term, and we will now proceed to prove the variance term.

\subsection{Proof of Theorem 3: Variance term}\label{sec: B.2}
To begin, let's also revisit the expressions for variance within our ridge regression framework at the regularization parameter $\lambda$:
\begin{align*}
    V_X(\widehat{B}_{\lambda};B)&=\frac{{\rm Tr}(\Sigma_{\epsilon})}{n}{\rm Tr}(\Sigma S_X(S_X+\lambda I)^{-2}) \\
    &={\rm Tr}(\Sigma_{\epsilon})\gamma \frac{\partial}{\partial\lambda}\{\frac{\lambda}{p}{\rm Tr}(\Sigma S_X(S_X+\lambda I)^{-2})\}.
\end{align*}
We define 
\begin{align*}
    L_n(\lambda):=\frac{1}{p}{\rm Tr}(\Sigma(I+r_n(-\lambda,0)\Sigma)^{-1}).
\end{align*}

Here, the $r_n$ is defined as previously mentioned. From equation (85) in~\citep{10.1214/21-AOS2133}, we can derive that
\begin{align*}
    |V_X(\widehat{B}_{\lambda};B)-{\rm Tr}(\Sigma_{\epsilon})\gamma \frac{\partial L_n}{\partial\lambda}(\lambda)|\leq \frac{C}{\lambda^2n^{(1-\epsilon)/2}}.
\end{align*}
Here the second component within the absolute value represents the expression of the prediction variance $\mathscr{V}(\lambda;\gamma, \Sigma_x)$ as defined in Definition \ref{def: 3}.\\

By combining the bias and variance terms, we have successfully completed the proof of Theorem \ref{thm: 3}. In the next stage, we will establish Theorem \ref{thm: 2} based on the findings in Theorem \ref{thm: 3}.

\section{Proof of Theorem 2}\label{sec: C}
The proof is derived from Theorem \ref{thm: 3}, which approximates min-norm regression with ridge regression using a small value of $\lambda$. Therefore, for the purpose of this proof, we will assume that $\lambda \leq 1$.

\subsection{Proof of Theorem 2: Bias term}\label{sec: C.1}
Recalling the expressions $S_X:=\frac{X^TX}{n}$ as defined in the proof of Theorem \ref{thm: 3} and in Lemma \ref{lem: 1}, and once more leveraging the result from Lemma \ref{lem: 2}, it becomes evident that there exists a vector $\beta$, for which $||\beta||_2=||B||_F$, and this $\beta$ satisfies:
\begin{align*}
    |B_X(\widehat{B}_{\lambda};B)-B_X(\widehat{B};B)|&=|\lambda^2{\rm Tr}(B^T(S_X+\lambda I)^{-1}\Sigma(S_X+\lambda I)^{-1}B) \\
    &-{\rm Tr}(B^T(S_XS_X^+-I)\Sigma(S_X^+S_X-I)B)| \\
    &\leq |\lambda^2{\rm Tr}(\beta^T(S_X+\lambda I)^{-1}\Sigma(S_X+\lambda I)^{-1}\beta) \\
    &-{\rm Tr}(\beta^T(S_XS_X^+-I)\Sigma(S_X^+S_X-I)\beta)| \\
    &\leq C(M) \lambda.
\end{align*}
The final step in this process can be derived from equation (104) in~\citep{10.1214/21-AOS2133}. To enhance readability and comprehension, we introduce the following definitions
\begin{align*}
    c_1:=c_0\frac{\sum_{i=1}^{p} \frac{s_i^2}{(1+c_0\gamma s_i)^2}}{\sum_{i=1}^{p}\frac{s_i}{(1+c_0\gamma s_i)^2}}.
\end{align*}

Taking into consideration the bias $\mathscr{B}(\gamma,\Sigma_x)$ defined in Definition \ref{def: 2} and the bias $\mathscr{B}(\lambda;\gamma, \Sigma_x)$ defined in Definition \ref{def: 3}. By following the reasoning presented in the proof of the variance term in Section B.1 of~\citep{10.1214/21-AOS2133}, we can ascertain that
\begin{align*}
    m_n(-\lambda)=&(1-\frac{1}{\gamma})\frac{1}{\lambda}+c_0+O(\lambda), \\
    m_{n, 1}(-\lambda)=&c_1+O(\lambda). \\
\end{align*}
Then we can express that
\begin{align*}
    |\mathscr{B}(\lambda;\gamma, \Sigma_x)-\mathscr{B}(\gamma,\Sigma_x)|&=|(1+\gamma c_1){\rm Tr}(B^T(I+c_0\gamma\Sigma)^{-2}\Sigma B) \\
    &-(1+\gamma c_1+\gamma O(\lambda)){\rm Tr}(B^T(I+c_0\gamma\Sigma+\gamma O(\lambda)\Sigma)^{-2}\Sigma B)|.
\end{align*}
Let
\begin{align*}
    f(x)=(1+\gamma c_1+x){\rm Tr}(B^T(I+c_0\gamma\Sigma+x\Sigma)^{-2}\Sigma B).
\end{align*}
Subsequently, in accordance with Lagrange's mean value theorem, there exists a value $\xi$ in the interval $[0, x]$ such that
\begin{align*}
    \frac{f(x)-f(0)}{x}=f^{'}(\xi).
\end{align*}
Thus, we can conclude that
\begin{align*}
    |\mathscr{B}(\lambda;\gamma, \Sigma_x)-\mathscr{B}(\gamma,\Sigma_x)|&=|f(\gamma O(\lambda))-f(0)|\\
    &=|\{{\rm Tr}(B^T(I+c_0\gamma\Sigma+\xi\Sigma)^{-2}\Sigma B) \\
    &-2(1+\gamma c_1+\xi){\rm Tr}(B^T(I+c_0\gamma\Sigma+\xi\Sigma)^{-3}\Sigma B)\}\gamma O(\lambda)| \\
    &\leq \{{\rm Tr}(B^T(I+c_0\gamma\Sigma+\xi\Sigma)^{-2}\Sigma B) \\
    &+2(1+\gamma c_1+\xi){\rm Tr}(B^T(I+c_0\gamma\Sigma+\xi\Sigma)^{-3}\Sigma B)\}\gamma O(\lambda) \\
    & \leq C(M)\lambda.
\end{align*}
In summary, we can derive that
\begin{align*}
    |B_X(\widehat{B};B)-\mathscr{B}(\gamma,\Sigma_x)|&\leq |B_X(\widehat{B}_{\lambda};B)-B_X(\widehat{B};B)|+|\mathscr{B}(\lambda;\gamma, \Sigma_x)-\mathscr{B}(\gamma,\Sigma_x)| \\
    &+|B_X(\widehat{B}_{\lambda};B)-\mathscr{B}(\lambda;\gamma, \Sigma_x)|\\
    &=C(M)(\lambda+\frac{||B||_F}{n^{(1-\epsilon)/2}\lambda}).
\end{align*}
In this context, we set $\lambda=n^{-1/4}$ and select $\epsilon$ to be sufficiently small. Consequently, we can attain the desired bound as stipulated in Theorem \ref{thm: 2}.

\subsection{Proof of Theorem 2: Variance term}\label{sec: C.2}
We denote the eigenvalue decomposition of $S_X$ as $S_X = U D_X U^T$, where $D_X \in \mathbb{R}^{p \times p}$ is a diagonal matrix, and $U \in \mathbb{R}^{p \times p}$ is an orthogonal matrix. 
Furthermore, we define $1_{D_X=0}$ as the diagonal matrix with the $(i, i)$-th entry equal to 1 if $(D_X)_{ii} = 0$ and equal to 0 otherwise. Subsequently, we can express the variance in Lemma 1 and the variance within our ridge regression framework as
\begin{align*}
    V_X(\widehat{B};B)&=\frac{{\rm Tr}(\Sigma_{\epsilon})}{n}{\rm Tr}(\Sigma UD_X^{-1}1_{D_X>0}U^T), \\
    V_X(\widehat{B}_{\lambda};B)&=\frac{{\rm Tr}(\Sigma_{\epsilon})}{n}{\rm Tr}(\Sigma UD_X(\lambda I+D_X)^{-2}U^T).
\end{align*}
Then from~\citep{10.1214/21-AOS2133} we can derive that
\begin{align*}
    |V_X(\widehat{B}_{\lambda};B)-V_X(\widehat{B};B)|\leq \frac{2\lambda M{\rm Tr}(\Sigma_{\epsilon})}{\sigma_{min}(X)^4/n^2}\leq C(M)\lambda.
\end{align*}
Taking into consideration the bias $\mathscr{V}(\gamma,\Sigma_x)$ defined in Definition \ref{def: 2} and the bias $\mathscr{V}(\lambda;\gamma, \Sigma_x)$ defined in Definition \ref{def: 3}, we can express that
\begin{align*}
    |\mathscr{V}(\lambda;\gamma, \Sigma_x)-\mathscr{V}(\gamma,\Sigma_x)|=&|{\rm Tr}(\Sigma_{\epsilon})\gamma \sum_{i=1}^{p}\frac{s_i^2(1-\gamma+\gamma\lambda^2 m_n^{'}(-\lambda))}{[\lambda+s_i(1-\gamma+\gamma\lambda m_n(-\lambda))]^2} \\
    &-{\rm Tr}(\Sigma_{\epsilon})\gamma c_0\frac{\sum_{i=1}^{p} \frac{s_i^2}{(1+c_0\gamma s_i)^2}}{\sum_{i=1}^{p}\frac{s_i}{(1+c_0\gamma s_i)^2}}|.
\end{align*}
Following the argument made in the proof of the variance term in Section B.1 of~\citep{10.1214/21-AOS2133}, we can conclude that
\begin{align*}
    |\mathscr{V}(\lambda;\gamma, \Sigma_x)-\mathscr{V}(\gamma,\Sigma_x)|\leq C(M)\lambda.
\end{align*}
To sum up, we can derive that
\begin{align*}
    |V_X(\widehat{B};B)-\mathscr{V}(\gamma,\Sigma_x)|&\leq |V_X(\widehat{B}_{\lambda};B)-V_X(\widehat{B};B)|+|\mathscr{V}(\lambda;\gamma, \Sigma_x)-\mathscr{V}(\gamma,\Sigma_x)| \\
    &+|V_X(\widehat{B}_{\lambda};B)-\mathscr{V}(\lambda;\gamma, \Sigma_x)|\\
    &=C(M)(\lambda+\frac{1}{\lambda^2n^{(1-\epsilon)/2}}).
\end{align*}
In this context, we also set $\lambda=n^{-1/4}$ and select $\epsilon$ to be sufficiently small. Consequently, we can attain the desired bound as stipulated in Theorem \ref{thm: 2}.\\

By merging the bias term and the variance term, we have successfully concluded the proof of Theorem \ref{thm: 2}.